\documentclass[letterpaper,journal]{IEEEtran}
\usepackage{amsmath,amsfonts,amssymb}
\usepackage{array}
\usepackage[caption=false,font=normalsize,labelfont=sf,textfont=sf]{subfig}
\usepackage{textcomp}
\usepackage{stfloats}
\usepackage{url}
\usepackage{graphicx}
\usepackage{cite}
\usepackage{algorithm}
\usepackage{algorithmic}
\usepackage[table]{xcolor}
\usepackage{multirow}
\usepackage{placeins}
\usepackage{xspace}
\definecolor{tableheader}{RGB}{220,220,220}
\definecolor{tablerow}{RGB}{245,245,245}
\definecolor{highlightcyan}{RGB}{218,238,247}
\DeclareRobustCommand{\ENCORE}{\textbf{ENCORE}\xspace}

\begin{document}

\bstctlcite{BSTcontrol}

\title{ENCORE: Event-Assisted Complementary Motion Refinement for Learned Video Compression}

\author{
Shuhan~Ye,
Hongbin~Yu,
Chenqi~Kong, Pingchuan~Ma,
Chong~Wang, Jun~Wan
and Qixin~Zhang
\IEEEcompsocitemizethanks{
\IEEEcompsocthanksitem
Shuhan Ye and Qixin Zhang are with the School of Computer Science,
Wuhan University, Wuhan, China (e-mail: shawnye618@gmail.com; qixin.zhang2026@gmail.com).
\IEEEcompsocthanksitem
Hongbin Yu and Chong Wang are with the School of Artificial Intelligence,
Ningbo University, Ningbo, China (e-mail: 2511100298@nbu.edu.cn; wangchong@nbu.edu.cn).
\IEEEcompsocthanksitem
Chenqi Kong is with the National University of Singapore, Singapore (e-mail: cqkong@nus.edu.sg).
\IEEEcompsocthanksitem
Pingchuan Ma is with the Zhejiang University of Technology, Hangzhou, China(e-mail: pma@zjut.edu.cn).
\IEEEcompsocthanksitem
Jun Wan is  with the School of Information Engineering, Zhongnan University of Economics and Law, Wuhan 430073, China,. (e-mail:junwan2014@whu.edu.cn).
\IEEEcompsocthanksitem
Corresponding authors: Jun Wan and Qixin Zhang.
}
}

\markboth{Journal of \LaTeX\ Class Files,~Vol.~14, No.~8, August~2021}%
{Ye: \ENCORE: Event-Assisted Complementary Motion Refinement for Learned Video Compression}

\maketitle

\begin{abstract}
Learned video compression relies on accurate temporal modeling to remove redundancy between adjacent frames. However, most existing codecs infer motion solely from discretely sampled RGB frames, making their estimates vulnerable to fast motion, blur, occlusion, weak texture, low illumination, and abrupt brightness changes. Event cameras asynchronously capture fine-grained intensity changes between RGB timestamps and therefore provide complementary evidence about inter-frame dynamics. We propose \ENCORE, an Event-Assisted Complementary Motion Refinement framework for learned video compression. \ENCORE first employs Complementary Motion Representation (CMR) to decompose aligned RGB--event features into common and modality-specific motion representations. Spatial Energy and Redundancy-Informed Calibration (SERIC) then identifies event-specific responses that are active and novel relative to RGB, suppresses weak or redundant evidence, and predicts a candidate flow correction. Finally, Energy-Aware Routing (EAR) determines where and how strongly the correction should refine the RGB flow. Events serve solely as an auxiliary modality for motion modeling, while RGB remains the only coding and reconstruction target. Experiments on BS-ERGB, HQ-EVFI, and CED demonstrate consistent gains across datasets and GOP lengths. On BS-ERGB, \ENCORE achieves up to $20.80\%$ PSNR-RGB and $22.14\%$ MS-SSIM-RGB BD-rate savings, while retaining clear improvements on the other two datasets. 
\end{abstract}


% ============================================================
% INTRODUCTION
% ============================================================

\section{Introduction}

Video data account for a large fraction of visual storage and network traffic, making efficient video compression essential for communication, streaming, surveillance, and edge visual systems. Learned video compression has emerged as an important direction for reducing bitrate under end-to-end rate-distortion optimization~\cite{lu2019dvc,lu2021e2e,habibian2019rdvae}. Instead of designing motion estimation, motion compensation, residual or conditional coding, and entropy modeling as isolated components, learned codecs optimize these modules jointly from data. Early neural codecs established predictive coding pipelines with learned motion and residual representations, while subsequent methods improved compression through hierarchical quality structures, recurrent and flexible-rate modeling, feature-space compensation, conditional coding, and richer temporal contexts~\cite{yang2020hlvc,yang2021recurrent,rippel2021elfvc,hu2021fvc,li2021dcvc,li2023diverse,gomes2023ecn,ho2022canfvc,mentzer2022nvcgan,sheng2023tcm,guo2024ecmf,sheng2025bidirectional}. Despite their architectural differences, these methods share a common principle: rate-distortion performance depends strongly on how accurately the codec exploits temporal redundancy between neighboring frames.

Most existing learned video codecs derive motion information and temporal context from the current RGB frame, previously reconstructed RGB frames, or previously extracted RGB features~\cite{lu2019dvc,lu2021e2e,hu2021fvc,li2021dcvc,sheng2023tcm}. This design is natural because RGB frames contain dense appearance and semantic information. However, an RGB video is only a discrete sampling of a continuously evolving dynamic scene, and the motion process between adjacent frames is not directly observed. When the scene contains fast motion, motion blur, occlusion, weak texture, low illumination, or abrupt brightness changes, inferring inter-frame dynamics from RGB appearance differences alone can be unreliable~\cite{agustsson2020ssf,li2023diverse}. In compression, inaccurate temporal modeling is not merely a reconstruction problem: it can degrade prediction quality, increase uncertainty in the motion and residual representations, and ultimately raise both motion and residual bitrates. These limitations suggest that RGB-only observations may be insufficient for reliable inter-frame dynamic modeling in challenging video compression scenarios.

Multimodal compression provides a useful perspective on this question and can be broadly categorized into two settings. The first is joint multimodal compression, which encodes and reconstructs multiple modalities together while exploiting their shared information to reduce the overall coding cost. Representative examples include stereo image compression, visible-infrared image and video compression, and RGB-D image compression~\cite{liu2019dsic,lu2022multimodality,zheng2024rgbd}. By aligning complementary observations and modeling cross-modal redundancy, these methods improve coding efficiency while preserving the information needed to reconstruct each modality. The second is auxiliary-modality compression, which uses an auxiliary modality to enhance the representation and coding of a designated primary modality without necessarily reconstructing the auxiliary signal as an additional output. Text-guided image compression methods, such as MMDN and TACO~\cite{jiang2023mmdn,lee2024taco}, demonstrate that semantic or complementary information from an auxiliary modality can strengthen RGB encoding and reconstruction. Despite these advances, multimodal video compression remains relatively underexplored, particularly in this auxiliary-modality setting. Methods that use auxiliary observations to alleviate dynamic ambiguities in RGB video compression, including those caused by fast motion, motion blur, occlusion, weak texture, and illumination variation, remain scarce.

Event cameras are particularly suitable as an auxiliary modality for RGB video. RGB frames provide dense color, texture, semantic content, and absolute intensity, whereas events asynchronously record pixel-level brightness changes with high temporal resolution, high dynamic range, and low latency~\cite{lichtsteiner2008dvs,gallego2022event}. In this sense, RGB frames describe a scene at selected timestamps, whereas events describe the changes between those timestamps. This complementarity has been demonstrated across RGB-event tasks, including video reconstruction~\cite{rebecq2019events,rebecq2021highspeed,zhu2022evsnn,gao2025eventreconstruction}, frame interpolation~\cite{tulyakov2021timelens,tulyakov2022timelenspp}, video super-resolution~\cite{jing2021frequency,lu2023eventvsr}, motion deblurring~\cite{sun2022efnet}, HDR reconstruction~\cite{yang2023hdr}, low-light enhancement~\cite{liang2024evlight,jiang2024eventlowlight}, and anomaly detection~\cite{qian2025ucfcrimedvs}. Related event-driven and spiking studies further investigate cross-modal knowledge transfer~\cite{wang2021evdistill}, knowledge distillation between artificial and spiking models~\cite{ye2025ckd}, event-efficient learning~\cite{ye2025pace}, motion-sensitive temporal modeling~\cite{ye2026fire}, and robustness to event-timing perturbations~\cite{yu2026retiming}. Nevertheless, these works primarily optimize reconstruction, enhancement, recognition, or representation learning objectives rather than the bitrate-distortion trade-off of RGB video compression. Moreover, event streams contain noise, modality-specific redundancy, and information irrelevant to the current RGB coding state; naive feature fusion therefore does not necessarily improve compression performance.

To address these limitations, we propose \ENCORE, an event-assisted learned video compression framework that exploits event streams as an auxiliary modality for RGB motion modeling. Unlike joint multimodal codecs that reconstruct all participating modalities, \ENCORE uses events to support temporal prediction while retaining RGB as the only coding and reconstruction target. Its Complementary Motion Representation (CMR) module decomposes aligned RGB--event features into common and modality-specific motion representations; its Spatial Energy and Redundancy-Informed Calibration (SERIC) module identifies event-specific responses that are active yet insufficiently represented by RGB; and its Energy-Aware Routing (EAR) module controls where and how strongly the resulting correction refines the RGB flow. Experiments show consistent improvements across datasets, capture conditions, and GOP lengths. The strongest result is obtained on BS-ERGB with GOP$=32$, where \ENCORE reduces the PSNR-RGB and MS-SSIM-RGB BD-rates by $20.80\%$ and $22.14\%$, respectively, relative to the RGB-only HyTIP backbone. Clear gains on the high-frame-rate HQ-EVFI dataset and CED, which is captured using a different event sensor, further demonstrate the effectiveness of event-guided motion refinement under varied capture conditions. Our contributions are threefold:
\begin{itemize}
    \item To the best of our knowledge, we are the first to use event data as an auxiliary modality for RGB video compression. Events improve temporal prediction, while RGB remains the only coding and reconstruction target.
    \item We develop a progressive complementary motion refinement framework that decomposes common and modality-specific motion, identifies active event evidence unexplained by RGB, and spatially routes the resulting flow correction. 
    % Theoretical analyses characterize the innovation, non-expansiveness, and bounded-routing properties of this design.
    \item We conduct comprehensive experiments across three RGB--event datasets and two GOP settings. Component ablations, qualitative comparisons, and complexity measurements demonstrate consistent coding gains with only a $7.3\%$ increase in total GFLOPs per P-frame.
\end{itemize}

% ============================================================
% RELATED WORK
% ============================================================

\section{Related Work}

\subsection{Learned Video Compression}

Learned video codecs jointly optimize motion estimation, compensation, content coding, and entropy modeling under an end-to-end rate-distortion objective. DVC~\cite{lu2019dvc} established a representative flow-based predictive pipeline, while Scale-Space Flow~\cite{agustsson2020ssf} improves robustness to uncertain motion and occlusion. FVC~\cite{hu2021fvc} moves compensation and residual coding into feature space, and heterogeneous deformable compensation combines motion cues with different spatial supports~\cite{wang2024hdc}. More recent codecs emphasize learned temporal context: DCVC~\cite{li2021dcvc} formulates inter-frame compression as conditional coding; Temporal Context Mining~\cite{sheng2023tcm} propagates multi-scale contexts; Motion Information Propagation strengthens the interaction between motion and frame coding~\cite{qi2023mip}; and DCVC-FM uses feature modulation and periodic refresh for variable-rate and long-sequence coding~\cite{li2024dcvcfm}. HyTIP~\cite{chen2025hytip} further combines decoded-frame propagation with compact implicit latent buffers. Despite these advances, their motion and context are inferred solely from RGB frames and can remain unreliable under fast motion, occlusion, blur, weak texture, or illumination variation.

\subsection{Multimodal Visual Compression}

Multimodal compression exploits statistical dependence among different observations of the same scene. Most existing methods follow a joint-coding setting in which every participating modality is both encoded and reconstructed. DSIC aligns stereo features through disparity-aware transforms~\cite{liu2019dsic}, while visible-infrared~\cite{lu2022multimodality} and RGB-D~\cite{zheng2024rgbd} codecs model shared and modality-specific information across heterogeneous sensors. In these systems, cross-modal alignment and redundancy modeling reduce the total rate required to preserve multiple outputs, but each reconstructed modality remains part of the coding objective and typically contributes to the transmitted representation. A related but distinct setting uses one modality only as auxiliary guidance for a designated primary target. MMDN~\cite{jiang2023mmdn} and TACO~\cite{lee2024taco}, for example, introduce textual priors into learned image compression without reconstructing text as an output. However, extending auxiliary guidance from semantic image information to video motion is nontrivial: an auxiliary temporal stream can contain sensor noise, information already captured by RGB, and responses unrelated to the current coding state. \ENCORE follows this auxiliary-modality setting but uses events specifically to refine RGB motion before compression, retaining RGB as the sole coding and reconstruction target.

\subsection{RGB-Event Vision Methods}

Event cameras asynchronously report pixel-level log-intensity changes and offer high temporal resolution, high dynamic range, and low latency, but do not provide dense color, absolute intensity, or complete texture~\cite{lichtsteiner2008dvs,gallego2022event}. They therefore complement RGB frames with fine-grained motion, dynamic boundaries, and illumination changes. RGB-event methods have exploited this complementarity for reconstruction~\cite{rebecq2019events,gao2025eventreconstruction}, interpolation~\cite{tulyakov2021timelens}, super-resolution~\cite{jing2021frequency,lu2023eventvsr}, deblurring~\cite{sun2022efnet}, HDR imaging~\cite{yang2023hdr}, low-light enhancement~\cite{liang2024evlight,jiang2024eventlowlight}, and anomaly detection~\cite{qian2025ucfcrimedvs}. Related event-driven learning studies explore cross-modal distillation~\cite{wang2021evdistill}, knowledge transfer between artificial and spiking networks~\cite{ye2025ckd}, efficient learning from dense events~\cite{ye2025pace}, and motion-sensitive temporal modeling~\cite{ye2026fire}. The sensitivity of event-driven models to event timing has also motivated explicit analyses of spike-retiming perturbations~\cite{yu2026retiming}. However, these works do not directly optimize the bitrate-distortion trade-off of RGB video coding. How to select complementary event evidence while suppressing noisy and redundant responses remains insufficiently explored.

\begin{figure*}[t]
    \centering
    \includegraphics[width=\linewidth]{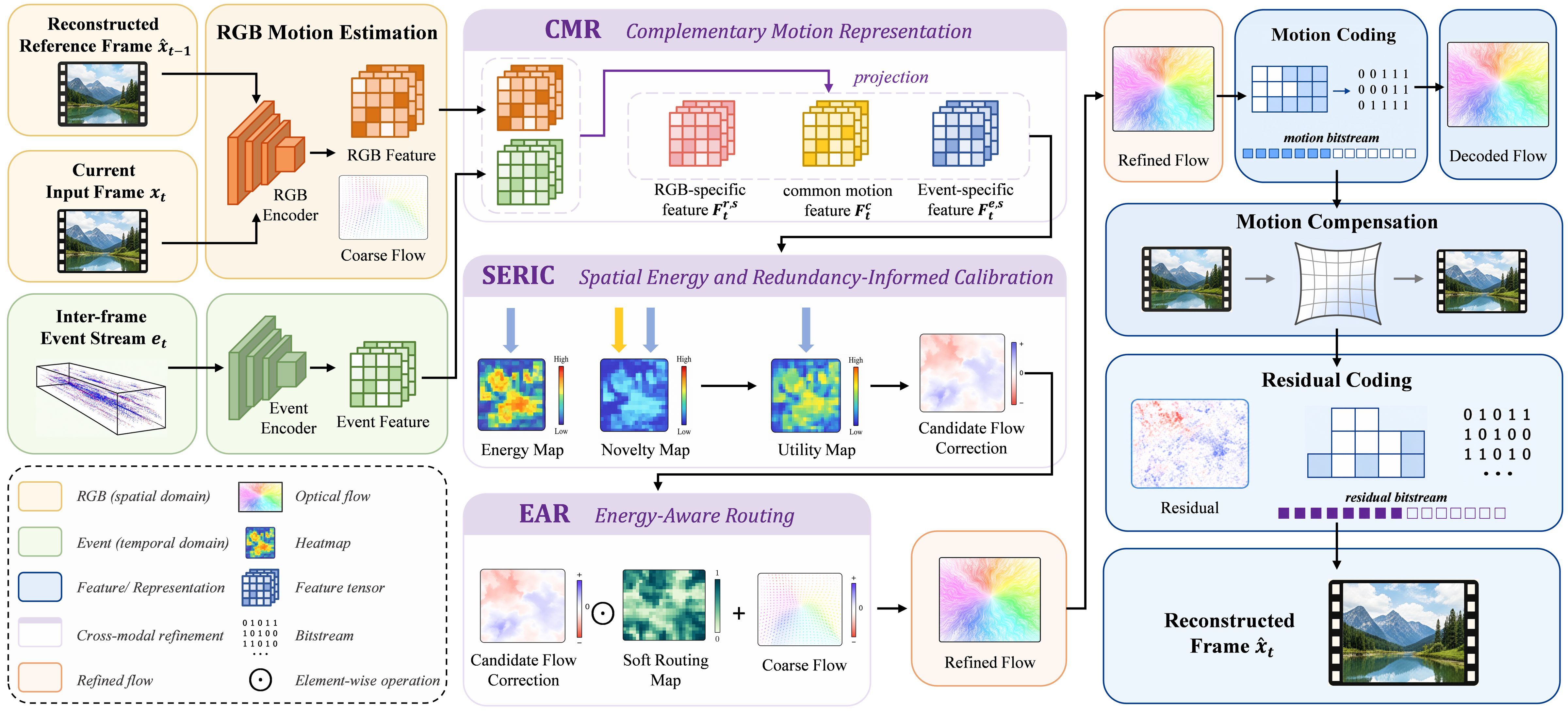}
    \caption{Overview of \ENCORE. CMR decomposes aligned RGB--event features, SERIC calibrates complementary event evidence to predict $\Delta\mathbf{v}_t$, and EAR routes this correction to obtain the refined flow $\widetilde{\mathbf{v}}_t$ for subsequent RGB coding.}
    \label{fig:overview}
\end{figure*}

% ============================================================
% METHOD
% ============================================================

\section{Methodology}
\label{sec:method}

\subsection{Problem Definition}

Let $\mathbf{x}_t\in[0,1]^{3\times H\times W}$ denote the current RGB frame after normalization from the conventional 8-bit range to $[0,1]$, and let $\hat{\mathbf{x}}_{t-1}$ denote the previously reconstructed frame in the same normalized domain. The events triggered between the timestamps of these two frames are collected as
\begin{equation}
    \mathbf{e}_t
    =
    \left\{
    \boldsymbol{\epsilon}_i=(x_i,y_i,\tau_i,p_i)
    \;\middle|\;
    \tau_i\in(\tau_{t-1},\tau_t]
    \right\},
    \label{eq:event_stream}
\end{equation}
where $(x_i,y_i)$, $\tau_i$, and $p_i\in\{-1,+1\}$ are the spatial location, timestamp, and polarity of event $\boldsymbol{\epsilon}_i$, respectively. In the considered setting, $\mathbf{e}_t$ is synchronized with the RGB pair and is available only to the encoder. The encoder produces the bitstream of frame $t$ as
\begin{equation}
    \mathbf{b}_t
    =
    \mathcal{E}_{\theta}
    \left(
    \mathbf{x}_t,\hat{\mathbf{x}}_{t-1},\mathbf{e}_t
    \right),
    \label{eq:enc_problem}
\end{equation}
where $\mathcal{E}_{\theta}$ denotes the encoding process with parameters $\theta$. The decoder reconstructs the current RGB frame according to
\begin{equation}
    \hat{\mathbf{x}}_t
    =
    \mathcal{D}_{\phi}
    \left(
    \mathbf{b}_t,\hat{\mathbf{x}}_{t-1}
    \right),
    \label{eq:dec_problem}
\end{equation}
where $\mathcal{D}_{\phi}$ is parameterized by $\phi$. Equations~\eqref{eq:enc_problem} and~\eqref{eq:dec_problem} make the asymmetric role of the two modalities explicit: RGB frames are the coding and reconstruction target, whereas events provide auxiliary observations for estimating the motion used by the RGB codec.

The codec is optimized for the rate--distortion performance of the reconstructed RGB video. For frame $t$, the basic objective is
\begin{equation}
    \mathcal{L}_{\mathrm{RD}}
    =
    R_t
    +
    \lambda
    D\!\left(\mathbf{x}_t,\hat{\mathbf{x}}_t\right),
    \label{eq:rd_loss}
\end{equation}
where $R_t$ is the expected coding rate during training (and the actual coded rate during evaluation), $D(\cdot,\cdot)$ measures RGB reconstruction distortion, and $\lambda$ controls the rate--distortion trade-off. The objective in Eq.~\eqref{eq:rd_loss} motivates \ENCORE: event observations are useful only when their motion cues reduce reconstruction uncertainty to improve the RGB rate--distortion balance.

\subsection{Overall Framework}

We propose \ENCORE, an \textbf{E}ve\textbf{N}t-Assisted \textbf{CO}mplementary Motion \textbf{RE}finement framework for learned video compression. We adopt the RGB video compression pipeline of HyTIP as the backbone~\cite{chen2025hytip}. \ENCORE is inserted into the encoder-side motion-estimation interface, before motion compression, and leaves the decoding interface unchanged. Given $\mathbf{x}_t$ and $\hat{\mathbf{x}}_{t-1}$, the RGB motion network $\mathcal{M}$ estimates an initial flow $\mathbf{v}_t^{r}$ together with its intermediate motion feature $\mathbf{F}_t^{r}$:
\begin{equation}
    \left(
    \mathbf{v}_t^{r},\mathbf{F}_t^{r}
    \right)
    =
    \mathcal{M}
    \left(
    \mathbf{x}_t,\hat{\mathbf{x}}_{t-1}
    \right).
    \label{eq:rgb_motion}
\end{equation}
The superscript $r$ identifies quantities estimated from RGB observations. In parallel, the event stem $\mathcal{T}_{\mathrm{motion}}$ converts the asynchronous stream into a motion-aligned representation
\begin{equation}
    \mathbf{F}_t^{e}
    =
    \mathcal{T}_{\mathrm{motion}}\!\left(\mathbf{e}_t\right),
    \label{eq:event_feature}
\end{equation}
whose spatial support and feature dimensionality are aligned with $\mathbf{F}_t^{r}$. Specifically, we partition the inter-frame events into $B=5$ temporal bins and accumulate positive and negative polarities separately, forming an event voxel grid $\mathbf{V}_t=\mathcal{V}_{B}(\mathbf{e}_t)\in\mathbb{R}^{2\times B\times H\times W}$. The event stem applies two lightweight $3\times3\times3$ convolutional blocks to $\mathbf{V}_t$, followed by temporal aggregation and a two-layer $2$D motion projection head. The spatial strides are selected to match the resolution of $\mathbf{F}_t^{r}$, while the final $1\times1$ projection matches its channel dimension. Equivalently,
\begin{equation}
    \mathbf{F}_t^{e}
    =
    \mathcal{P}_{\mathrm{motion}}
    \left(
    \mathcal{A}_{\tau}
    \left(
    \mathcal{E}_{3\mathrm{D}}
    \left(\mathbf{V}_t\right)
    \right)
    \right),
    \qquad
    \mathbf{V}_t=\mathcal{V}_{B}\left(\mathbf{e}_t\right),
    \label{eq:event_stem}
\end{equation}
where $\mathcal{E}_{3\mathrm{D}}$ denotes the spatiotemporal convolutional encoder, $\mathcal{A}_{\tau}$ denotes temporal aggregation, and $\mathcal{P}_{\mathrm{motion}}$ denotes the $2$D motion projection head. This design preserves the temporal ordering and polarity of inter-frame changes instead of collapsing them into a single event frame. Spatial and channel matching alone, however, do not guarantee motion semantics. During training, the auxiliary flow projection in Eq.~\eqref{eq:event_motion_projection} and the reliability-weighted alignment loss in Eq.~\eqref{eq:event_motion_loss} encourage $\mathbf{F}_t^{e}$ to predict displacement compatible with reliable RGB flow, while leaving unreliable RGB regions free to retain event-specific evidence. Thus, the stem maps event dynamics into a motion-oriented auxiliary representation rather than merely a spatially aligned feature tensor.

\ENCORE refines $\mathbf{v}_t^{r}$ through three consecutive stages. First, the Complementary Motion Representation (CMR) module decomposes the aligned features into cross-modal common motion $\mathbf{F}_t^{c}$, RGB-specific motion $\mathbf{F}_t^{r,s}$, and event-specific motion $\mathbf{F}_t^{e,s}$ (Section~\ref{sec:cmr}):
\begin{equation}
    \left(
    \mathbf{F}_t^{c},\mathbf{F}_t^{r,s},\mathbf{F}_t^{e,s}
    \right)
    =
    \mathcal{C}
    \left(
    \mathbf{F}_t^{r},\mathbf{F}_t^{e}
    \right).
    \label{eq:cmr_overall}
\end{equation}
Second, the Spatial Energy and Redundancy-Informed Calibration (SERIC) module evaluates the local activity and cross-modal novelty of $\mathbf{F}_t^{e,s}$ to identify the portion that provides complementary evidence beyond RGB (Section~\ref{sec:seric}). It suppresses weak or redundant responses and predicts a candidate flow correction $\Delta\mathbf{v}_t$. Third, the Energy-Aware Routing (EAR) module estimates a spatial routing map $\mathbf{G}_t$ that controls where and to what extent this correction should modify the RGB estimate (Section~\ref{sec:ear}). The resulting flow is
\begin{equation}
    \widetilde{\mathbf{v}}_t
    =
    \mathbf{v}_t^{r}
    +
    \mathbf{G}_t\odot\Delta\mathbf{v}_t,
    \label{eq:refined_flow_overall}
\end{equation}
where $\odot$ denotes element-wise multiplication. The refined flow $\widetilde{\mathbf{v}}_t$ replaces $\mathbf{v}_t^{r}$ in the subsequent motion coding, motion compensation, and residual coding stages; all remaining components of the RGB compression pipeline retain their original interfaces. Thus, CMR separates common from modality-specific motion, SERIC determines which event-specific responses provide useful evidence beyond RGB, and EAR determines where and how strongly the correction is injected.

\begin{algorithm}[t]
\caption{Encoder-Side \ENCORE Coding for Frame $t$}
\label{alg:encore}
\begin{algorithmic}[1]
\REQUIRE $\mathbf{x}_t$, $\hat{\mathbf{x}}_{t-1}$, inter-frame events $\mathbf{e}_t$
\ENSURE RGB bitstream $\mathbf{b}_t$, reconstruction $\hat{\mathbf{x}}_t$
\STATE \textbf{Motion evidence extraction:} estimate RGB motion
\STATE \quad $(\mathbf{v}_t^r,\mathbf{F}_t^r)\leftarrow
\mathcal{M}(\mathbf{x}_t,\hat{\mathbf{x}}_{t-1})$.
\STATE Map temporally ordered events to a motion-aligned representation
\STATE \quad $\mathbf{F}_t^e\leftarrow\mathcal{T}_{\mathrm{motion}}(\mathbf{e}_t)$.
\STATE \textbf{CMR decomposition:} separate $[\mathbf{F}_t^r,\mathbf{F}_t^e]$
\STATE \quad into common $\mathbf{F}_t^c$, RGB-specific $\mathbf{F}_t^{r,s}$,
\STATE \quad and event-specific candidate motion $\mathbf{F}_t^{e,s}$.
\STATE \textbf{SERIC calibration:} measure local event activity $\mathbf{E}_t$
\STATE \quad and RGB--event novelty $\mathbf{N}_t$.
\STATE Infer utility $\mathbf{U}_t\leftarrow
\sigma(\mathcal{G}([\mathbf{E}_t,\mathbf{N}_t]))$ and calibrate
\STATE \quad $\widetilde{\mathbf{F}}_t^{e,s}\leftarrow
\mathbf{U}_t\odot\mathbf{F}_t^{e,s}$.
\STATE Predict $\Delta\mathbf{v}_t\leftarrow
\mathcal{R}([\mathbf{F}_t^c,\widetilde{\mathbf{F}}_t^{e,s}])$.
\STATE \textbf{EAR routing:} infer $\mathbf{G}_t$ and update supported regions,
\STATE \quad $\widetilde{\mathbf{v}}_t\leftarrow
\mathbf{v}_t^r+\mathbf{G}_t\odot\Delta\mathbf{v}_t$.
\STATE \textbf{Standard RGB coding:} encode motion as
\STATE \quad $(\mathbf{b}_t^m,\hat{\mathbf{v}}_t,R_t^m)\leftarrow
\mathcal{K}_m(\widetilde{\mathbf{v}}_t)$.
\STATE Motion-compensate
$\bar{\mathbf{x}}_t\leftarrow
\mathcal{W}(\hat{\mathbf{x}}_{t-1},\hat{\mathbf{v}}_t)$.
\STATE Encode the remaining RGB information as
\STATE \quad $(\mathbf{b}_t^y,\hat{\mathbf{x}}_t,R_t^y)\leftarrow
\mathcal{K}_y(\mathbf{x}_t,\bar{\mathbf{x}}_t)$ and set
\STATE \quad $\mathbf{b}_t\leftarrow[\mathbf{b}_t^m,\mathbf{b}_t^y]$.
\RETURN $\mathbf{b}_t$ and the reconstructed RGB frame $\hat{\mathbf{x}}_t$
\end{algorithmic}
\end{algorithm}

\subsection{Complementary Motion Representation}
\label{sec:cmr}

RGB frames and events observe the same scene dynamics through different sensing mechanisms. Their motion features therefore contain common structures, such as moving boundaries and dominant displacement patterns, but are not interchangeable. RGB features retain dense appearance support and may miss rapid inter-frame changes, whereas event features directly describe brightness transitions but lack complete appearance information. Naive feature concatenation does not distinguish these roles: common responses can be represented repeatedly, while genuinely complementary event cues can be diluted. More importantly, event observations may contain sparse sensor noise, scattered activity, and modality-specific responses unrelated to the current RGB coding state. Passing all of these responses directly to the flow-refinement network can introduce harmful perturbations, so that RGB+E Concat may even underperform the RGB-only baseline. CMR addresses this issue by organizing the two feature streams into a cross-modal common representation, an RGB-specific representation, and an event-specific candidate representation. The subsequent SERIC stage determines which parts of this event-specific candidate provide genuinely complementary evidence beyond RGB.

Given $\mathbf{F}_t^{r}$ and $\mathbf{F}_t^{e}$, the common branch jointly extracts motion structures supported by both modalities:
\begin{equation}
    \mathbf{F}_t^{c}
    =
    \mathcal{C}_{c}
    \left(
    \left[\mathbf{F}_t^{r},\mathbf{F}_t^{e}\right]
    \right),
    \label{eq:cmr_common}
\end{equation}
where $[\cdot,\cdot]$ denotes channel-wise concatenation and $\mathcal{C}_{c}$ is a learnable common-motion projection that receives both modalities. Two separate modality-oriented projections, $\mathcal{C}_{r}$ and $\mathcal{C}_{e}$, retain the information needed to explain each input:
\begin{equation}
    \mathbf{F}_t^{r,s}
    =
    \mathcal{C}_{r}\!\left(\mathbf{F}_t^{r}\right),
    \qquad
    \mathbf{F}_t^{e,s}
    =
    \mathcal{C}_{e}\!\left(\mathbf{F}_t^{e}\right).
    \label{eq:cmr_specific}
\end{equation}
Thus, $\mathcal{C}_{c}$, $\mathcal{C}_{r}$, and $\mathcal{C}_{e}$ are lightweight learnable feature transforms for the common, RGB-specific, and event-specific branches, respectively, and their outputs share the same channel dimension. Here, $\mathbf{F}_t^{r,s}$ captures RGB-specific support and participates in the CMR reconstruction and separation constraints, whereas $\mathbf{F}_t^{e,s}$ preserves event-derived dynamics that may provide evidence beyond the RGB representation. The latter remains an explicit input to SERIC, which evaluates its activity and novelty before flow refinement.

The decomposition must satisfy two requirements. First, the common and modality-oriented components should jointly preserve the information carried by the original features. We reconstruct the two inputs as
\begin{equation}
    \hat{\mathbf{F}}_t^{r}
    =
    \mathcal{P}_{r}
    \left(
    \left[\mathbf{F}_t^{c},\mathbf{F}_t^{r,s}\right]
    \right),
    \qquad
    \hat{\mathbf{F}}_t^{e}
    =
    \mathcal{P}_{e}
    \left(
    \left[\mathbf{F}_t^{c},\mathbf{F}_t^{e,s}\right]
    \right),
    \label{eq:cmr_reconstruction}
\end{equation}
where $\mathcal{P}_{r}$ and $\mathcal{P}_{e}$ are the RGB and event feature reconstruction projections, respectively. We define the reconstruction loss
\begin{equation}
    \mathcal{L}_{r}
    =
    \left\|
    \hat{\mathbf{F}}_t^{r}-\mathbf{F}_t^{r}
    \right\|_{1}
    +
    \left\|
    \hat{\mathbf{F}}_t^{e}-\mathbf{F}_t^{e}
    \right\|_{1}.
    \label{eq:cmr_rec_loss}
\end{equation}
This constraint prevents the decomposition from collapsing to uninformative feature partitions. Second, the common representation should not duplicate the information assigned to either modality-oriented branch. We therefore use
\begin{equation}
    \mathcal{L}_{s}
    =
    \operatorname{sim}
    \left(
    \mathbf{F}_t^{c},\mathbf{F}_t^{r,s}
    \right)
    +
    \operatorname{sim}
    \left(
    \mathbf{F}_t^{c},\mathbf{F}_t^{e,s}
    \right)
    +
    \operatorname{sim}
    \left(
    \mathbf{F}_t^{r,s},\mathbf{F}_t^{e,s}
    \right).
    \label{eq:cmr_sep_loss}
\end{equation}
For a feature tensor $\mathbf{A}$, let $\mathbf{z}(\mathbf{A})$ denote its channel vector after global average pooling and $L_2$ normalization. We implement
\begin{equation}
    \operatorname{sim}(\mathbf{A},\mathbf{B})
    =
    \frac{1}{N_b}
    \sum_{n=1}^{N_b}
    \left(
    \mathbf{z}(\mathbf{A})_n^{\mathsf T}
    \mathbf{z}(\mathbf{B})_n
    \right)^2,
    \label{eq:cmr_similarity}
\end{equation}
where $N_b$ is the batch size. Thus, $\operatorname{sim}(\cdot,\cdot)$ is the batch-averaged squared cosine similarity. Squaring penalizes both positive and negative correlation and makes orthogonality its minimum. Equation~\eqref{eq:cmr_sep_loss} applies this penalty to all three pairs of common, RGB-specific, and event-specific representations. Combined with the reconstruction loss $\mathcal{L}_{r}$, it discourages redundant feature partitions while preserving $\mathbf{F}_t^{e,s}$ as candidate motion evidence for subsequent novelty calibration.

\subsection{Spatial Energy and Redundancy-Informed Calibration}
\label{sec:seric}

The event-specific candidate representation can still contain responses that are unhelpful for compression. Sensor activity may be weak or spatially scattered, and some event structures may already be sufficiently captured by the RGB motion feature. Applying all responses with equal strength would therefore introduce unnecessary perturbations into the motion estimate. SERIC identifies complementary evidence within $\mathbf{F}_t^{e,s}$ by jointly considering its local response energy and its novelty relative to the RGB branch.

We first construct a spatial activity map from the squared magnitude of the event-specific candidate feature:
\begin{equation}
    \mathbf{E}_t
    =
    \mathcal{P}
    \left(
    \left|\mathbf{F}_t^{e,s}\right|^2
    \right),
    \label{eq:event_energy}
\end{equation}
where $\mathcal{P}(\cdot)$ denotes local spatial aggregation over feature channels and neighboring positions. The map $\mathbf{E}_t$ is an energy-inspired activity measurement rather than a physical energy quantity: a large value indicates that the event branch responds strongly in the corresponding region. Activity alone, however, cannot determine whether the response contributes information beyond the RGB estimate. We project the RGB feature into the event-specific feature space using $\psi(\cdot)$ and compute
\begin{equation}
    \mathbf{N}_t
    =
    \mathcal{P}
    \left(
    \left|
    \mathbf{F}_t^{e,s}
    -
    \psi\!\left(\mathbf{F}_t^{r}\right)
    \right|
    \right),
    \label{eq:event_novelty}
\end{equation}
where $\mathbf{N}_t$ measures cross-modal novelty. A high response indicates that the event feature is poorly explained by the aligned RGB motion feature and is therefore more likely to provide complementary dynamic evidence.

This discrepancy also admits an innovation interpretation. Under a squared-error idealization, the optimal RGB-to-event predictor is the conditional mean $\mathbb{E}[\mathbf{F}_t^{e,s}\mid\mathbf{F}_t^r]$, and its residual is orthogonal to every square-integrable function of the RGB feature. Thus, Eq.~\eqref{eq:event_novelty} serves as a robust local surrogate for the magnitude of event information unexplained by RGB; it does not assume that the learned $\psi$ exactly attains the conditional mean. Appendix~D-A provides the formal result.

The activity and novelty maps describe different failure cases and are combined to estimate event utility:
\begin{equation}
    \mathbf{U}_t
    =
    \sigma
    \left(
    \mathcal{G}
    \left(
    \left[\mathbf{E}_t,\mathbf{N}_t\right]
    \right)
    \right),
    \label{eq:event_utility}
\end{equation}
where $\mathcal{G}$ is a learnable calibration function and $\sigma(\cdot)$ is the sigmoid function. The resulting map $\mathbf{U}_t\in[0,1]$ is trained to attenuate inactive or redundant responses while retaining regions that exhibit meaningful event activity and a discrepancy from the RGB representation. The calibrated event feature is
\begin{equation}
    \widetilde{\mathbf{F}}_t^{e,s}
    =
    \mathbf{U}_t\odot\mathbf{F}_t^{e,s}.
    \label{eq:calibrated_event}
\end{equation}
Broadcasting is applied over feature channels when $\mathbf{U}_t$ is spatial. This multiplicative form retains the selected event responses while explicitly controlling their amplitude.

Because every element of $\mathbf{U}_t$ lies in $[0,1]$, the calibration is non-expansive under any standard $\ell_p$ norm: it can preserve or attenuate the event-specific response but cannot amplify its norm. Under a Lipschitz correction predictor, this property also bounds the output perturbation attributable to the event branch. The precise statement is given in Appendix~D-B.

SERIC combines the common motion structure with calibrated event evidence to predict a candidate flow correction:
\begin{equation}
    \Delta\mathbf{v}_t
    =
    \mathcal{R}
    \left(
    \left[
    \mathbf{F}_t^{c},
    \widetilde{\mathbf{F}}_t^{e,s}
    \right]
    \right).
    \label{eq:candidate_correction}
\end{equation}
The common feature anchors the prediction to motion structures supported by both sensors, whereas $\widetilde{\mathbf{F}}_t^{e,s}$ supplies calibrated changes that are insufficiently represented by the RGB branch. The output $\Delta\mathbf{v}_t$ has the same two-channel spatial layout as $\mathbf{v}_t^{r}$ after resolution alignment. It represents a candidate residual update rather than a replacement flow, which allows the original RGB estimate to remain the reference for the final routing stage.

\begin{figure*}[ht]
    \centering
    \includegraphics[width=\textwidth]{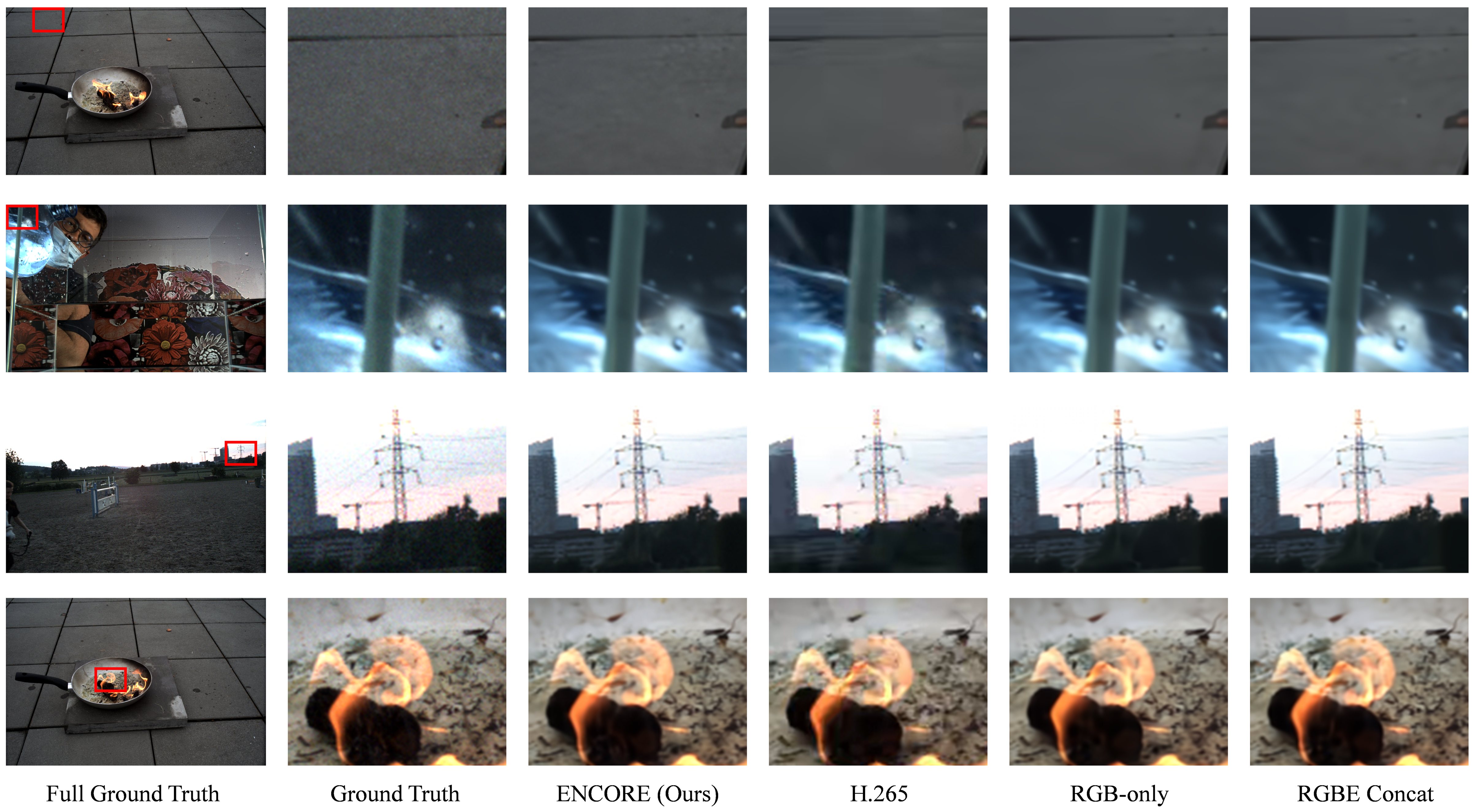}
    \caption{Qualitative comparison on BS-ERGB at comparable bitrates. From top to bottom, the examples are from \textit{``fire\_02''}, \textit{``may29\_water\_tank\_pouring\_02''}, \textit{``horse\_20''}, and \textit{``fire\_02''}, respectively. From left to right, the columns show the full ground-truth frame, ground-truth crop, \ENCORE, H.265/HM, RGB-only, and RGB+E Concat. Red boxes indicate the enlarged regions.}
    \label{fig:qualitative}
\end{figure*}

\subsection{Energy-Aware Routing}
\label{sec:ear}

Calibration determines which event features are informative, but it does not guarantee that the predicted flow correction is beneficial at every spatial location. In regions where the initial RGB flow is already reliable, an unnecessary update may degrade motion compensation and increase both motion and residual coding costs. EAR therefore converts the evidence accumulated by CMR and SERIC into a bounded routing decision that controls the spatial injection of $\Delta\mathbf{v}_t$.

Specifically, the routing network $\mathcal{Q}$ receives the local activity $\mathbf{E}_t$, cross-modal novelty $\mathbf{N}_t$, the magnitude of the initial RGB flow, and the magnitude of the candidate correction. We denote the per-pixel $\ell_2$ magnitude of a two-channel flow by $\operatorname{mag}(\mathbf{v})_{i,j}=\sqrt{v_{x,i,j}^{2}+v_{y,i,j}^{2}}$. After aligning these maps to a common spatial resolution, EAR predicts
\begin{equation}
    \mathbf{G}_t
    =
    \sigma
    \left(
    \mathcal{Q}
    \left(
    \left[
    \mathbf{E}_t,
    \mathbf{N}_t,
    \operatorname{mag}\!\left(\mathbf{v}_t^{r}\right),
    \operatorname{mag}\!\left(\Delta\mathbf{v}_t\right)
    \right]
    \right)
    \right),
    \label{eq:routing_map}
\end{equation}
where $H_v$ and $W_v$ denote the spatial dimensions of the flow, and $\mathbf{G}_t\in[0,1]^{H_v\times W_v}$ is the corresponding routing map. The map is shared by the horizontal and vertical components of the correction through channel broadcasting. Event activity and novelty indicate whether complementary evidence exists, while the two motion magnitudes provide the local context needed to regulate the scale of the proposed update.

The final refined flow is obtained through residual routing:
\begin{equation}
    \widetilde{\mathbf{v}}_t
    =
    \mathbf{v}_t^{r}
    +
    \mathbf{G}_t\odot\Delta\mathbf{v}_t.
    \label{eq:routed_refinement}
\end{equation}
When the event response is weak, redundant with RGB, or associated with an unsuitable correction, the routing weight can approach zero and Eq.~\eqref{eq:routed_refinement} recovers the original RGB estimate locally. In regions with strong complementary dynamics, a larger weight admits more of the candidate correction. The bounded residual form also prevents \ENCORE from discarding the motion structure already learned by the RGB branch. In this way, the three modules form a progressive refinement process: CMR exposes modality-specific event candidates, SERIC identifies and calibrates their complementary utility, and EAR translates that utility into a spatially controlled modification of the flow used for compression.

The bounded map also guarantees $\|\mathbf{G}_t\odot\Delta\mathbf{v}_t\|_p\leq\|\Delta\mathbf{v}_t\|_p$, so routing cannot enlarge the candidate update. Moreover, at each location, the oracle scalar weight in $[0,1]$ minimizes a convex quadratic flow error and is no worse than either rejecting the correction or applying it in full. EAR is learned as a spatial approximation to this oracle rule; no claim is made that it always reaches the oracle solution. Appendix~D-C gives the derivation and stability result. Algorithm~\ref{alg:encore} summarizes how the three modules are embedded into the complete encoder-side coding process for frame $t$.

\subsection{Compression and Optimization}

After refinement, $\widetilde{\mathbf{v}}_t$ enters the standard RGB motion codec, denoted by $\mathcal{K}_{m}$:
\begin{equation}
    \left(
    \mathbf{b}_t^m,\hat{\mathbf{v}}_t,R_t^{m}
    \right)
    =
    \mathcal{K}_{m}
    \left(
    \widetilde{\mathbf{v}}_t
    \right).
    \label{eq:motion_compression}
\end{equation}
The decoded flow warps the previous reconstruction through differentiable warping $\mathcal{W}$, and the content codec $\mathcal{K}_{y}$ encodes the remaining information:
\begin{equation}
    \bar{\mathbf{x}}_t
    =
    \mathcal{W}
    \left(
    \hat{\mathbf{x}}_{t-1},\hat{\mathbf{v}}_t
    \right),
    \label{eq:motion_compensation}
\end{equation}
\begin{equation}
    \left(
    \mathbf{b}_t^y,\hat{\mathbf{x}}_t,R_t^{y}
    \right)
    =
    \mathcal{K}_{y}
    \left(
    \mathbf{x}_t,\bar{\mathbf{x}}_t
    \right).
    \label{eq:residual_compression}
\end{equation}
Thus, the RGB rate is
\begin{equation}
    R_t
    =
    R_t^{m}+R_t^{y},
    \label{eq:total_rate}
\end{equation}
where $R_t^{m}$ and $R_t^{y}$ include the motion and content side information, respectively. The event modality influences RGB coding through the refined motion estimate $\widetilde{\mathbf{v}}_t$.

Training combines the RGB rate--distortion objective with the CMR losses $\mathcal{L}_{r}$ and $\mathcal{L}_{s}$. To make the event representation motion-aware, a training-only auxiliary head predicts a coarse flow:
\begin{equation}
    \mathbf{v}_t^{e}
    =
    \mathcal{H}_{\mathrm{aux}}\!\left(\mathbf{F}_t^{e}\right).
    \label{eq:event_motion_projection}
\end{equation}
We compute the per-pixel photometric error between $\mathbf{x}_t$ and the prediction obtained by warping $\hat{\mathbf{x}}_{t-1}$ with $\mathbf{v}_t^r$. A monotonically decreasing normalization of this error defines the reliability mask $\mathbf{M}_t\in[0,1]^{H_v\times W_v}$, so that well-aligned pixels receive larger weights and poorly aligned pixels receive smaller weights. The motion-alignment loss is
\begin{equation}
    \mathcal{L}_{m}
    =
    \left\|
    \mathbf{M}_t\odot
    \left(
    \mathbf{v}_t^{e}
    -
    \operatorname{sg}\!\left(\mathbf{v}_t^{r}\right)
    \right)
    \right\|_1,
    \label{eq:event_motion_loss}
\end{equation}
where $\operatorname{sg}(\cdot)$ denotes stop-gradient. This loss distills reliable RGB displacement into the event stem without constraining regions in which RGB flow is unreliable. The auxiliary head is discarded at inference. We additionally supervise the refined pre-quantization flow with
\begin{equation}
    \mathcal{L}_{w}
    =
    \left\|
    \mathbf{x}_t
    -
    \mathcal{W}
    \left(
    \hat{\mathbf{x}}_{t-1},\widetilde{\mathbf{v}}_t
    \right)
    \right\|_{1}.
    \label{eq:warp_loss}
\end{equation}
The complete training objective is
\begin{equation}
    \mathcal{L}
    =
    \mathcal{L}_{\mathrm{RD}}
    +
    \lambda_{r}\mathcal{L}_{r}
    +
    \lambda_{s}\mathcal{L}_{s}
    +
    \lambda_{m}\mathcal{L}_{m}
    +
    \lambda_{w}\mathcal{L}_{w},
    \label{eq:full_loss}
\end{equation}
where $\lambda_{r}$, $\lambda_{s}$, $\lambda_{m}$, and $\lambda_{w}$ weight the auxiliary terms. The rate--distortion term remains primary, and all auxiliary losses are used only during training without changing the RGB reconstruction target. Their values are provided in Appendix~C.

\begin{table*}[t]
\centering
\caption{BD-rate (\%) on the three RGB--event datasets for learned codecs trained with GOP$=3$ and evaluated with GOP$=8$. RGB-only (HyTIP) is the anchor; negative values indicate bitrate savings.}
\label{tab:main_results}
\scriptsize
\setlength{\tabcolsep}{5.5pt}
\renewcommand{\arraystretch}{1.12}
\resizebox{\textwidth}{!}{%
\begin{tabular}{l c || cc | cc | cc}
\hline
\rowcolor{tableheader}
&
&
\multicolumn{2}{c|}{\textbf{BS-ERGB}} &
\multicolumn{2}{c|}{\textbf{HQ-EVFI}} &
\multicolumn{2}{c}{\textbf{CED}} \\
\rowcolor{tableheader}
\multirow{-2}{*}{\textbf{Method}} & \multirow{-2}{*}{\textbf{Input}} & PSNR-RGB & MS-SSIM-RGB & PSNR-RGB & MS-SSIM-RGB & PSNR-RGB & MS-SSIM-RGB \\
\hline\hline
RGB-only (HyTIP~\cite{chen2025hytip}) & RGB & 0.0 & 0.0 & 0.0 & 0.0 & 0.0 & 0.0 \\
\rowcolor{tablerow}
RGB+E Concat & RGB+E & +4.36 & +5.94 & +7.90 & +4.57 & +11.23 & +9.03 \\
\rowcolor{highlightcyan}
\textbf{\ENCORE{} (Ours)} & \textbf{RGB+E} & \textbf{-19.31} & \textbf{-16.73} & \textbf{-9.20} & \textbf{-9.61} & \textbf{-5.63} & \textbf{-6.49} \\
\hline
\end{tabular}%
}
\end{table*}

\begin{table*}[t]
\centering
\caption{Long-GOP BD-rate (\%) for learned codecs initialized from GOP$=3$, fine-tuned with GOP$=8$, and evaluated with GOP$=32$. RGB-only (HyTIP) is the anchor; negative values indicate bitrate savings.}
\label{tab:long_gop_results}
\scriptsize
\setlength{\tabcolsep}{5.5pt}
\renewcommand{\arraystretch}{1.12}
\resizebox{\textwidth}{!}{%
\begin{tabular}{l c || cc | cc | cc}
\hline
\rowcolor{tableheader}
&
&
\multicolumn{2}{c|}{\textbf{BS-ERGB}} &
\multicolumn{2}{c|}{\textbf{HQ-EVFI}} &
\multicolumn{2}{c}{\textbf{CED}} \\
\rowcolor{tableheader}
\multirow{-2}{*}{\textbf{Method}} & \multirow{-2}{*}{\textbf{Input}} & PSNR-RGB & MS-SSIM-RGB & PSNR-RGB & MS-SSIM-RGB & PSNR-RGB & MS-SSIM-RGB \\
\hline\hline
RGB-only (HyTIP~\cite{chen2025hytip}) & RGB & 0.0 & 0.0 & 0.0 & 0.0 & 0.0 & 0.0 \\
\rowcolor{tablerow}
RGB+E Concat & RGB+E & -4.44 & +3.90 & +10.69 & +4.66 & +6.66 & -2.79 \\
\rowcolor{highlightcyan}
\textbf{\ENCORE{} (Ours)} & \textbf{RGB+E} & \textbf{-20.80} & \textbf{-22.14} & \textbf{-9.60} & \textbf{-11.90} & \textbf{-6.43} & \textbf{-4.35} \\
\hline
\end{tabular}%
}
\end{table*}

% ============================================================
% EXPERIMENTS
% ============================================================

\section{Experiments}

\subsection{Datasets}

We evaluate on BS-ERGB~\cite{tulyakov2022timelenspp}, HQ-EVFI~\cite{ma2024timelensxl}, and CED~\cite{scheerlinck2019ced}, which provide synchronized RGB frames and event streams under complementary scene content and sensor characteristics. BS-ERGB is the primary dataset for training, testing, and ablation; HQ-EVFI evaluates performance on high-frame-rate RGB--event data; and CED provides an additional evaluation using a different event sensor. We train dataset-specific models for all three datasets using the same training and evaluation protocol; no \ENCORE model trained on BS-ERGB is transferred to HQ-EVFI or CED. Detailed dataset characteristics are provided in Appendix~A.

\subsection{Evaluation Metrics}

Following HyTIP~\cite{chen2025hytip}, we report RGB bitrate in bits per pixel (bpp):
\begin{equation}
    \mathrm{bpp}
    =
    \frac{B_{\mathrm{total}}}{N H W},
\end{equation}
where $B_{\mathrm{total}}$ denotes the total coded RGB bits for $N$ frames of size $H\times W$. We assess RGB reconstruction quality using PSNR-RGB and MS-SSIM-RGB, and summarize rate--distortion differences using BD-rate over the common quality range. The RGB-only HyTIP backbone is the common anchor; negative BD-rate indicates bitrate savings. Detailed metric definitions and evaluation conventions are provided in Appendix~B.

\subsection{Implementation Details}

We initialize the RGB branch from pretrained HyTIP and use two-stage optimization: the event-assisted modules are first trained with the RGB backbone frozen, after which the complete codec is jointly fine-tuned. Standard models are trained with GOP$=3$ and evaluated with GOP$=8$; the long-GOP models are further fine-tuned with GOP$=8$ and tested with GOP$=32$. Within each dataset, all learned variants use identical data splits, preprocessing, augmentation, training schedules, GOP protocols, and operating points. Complexity and runtime are measured on NVIDIA RTX PRO 6000 GPUs. Appendix~C provides the complete optimization hyperparameters, data-split, augmentation, resolution, and processing details.

\begin{figure*}[ht]
    \centering
    \includegraphics[width=\textwidth]{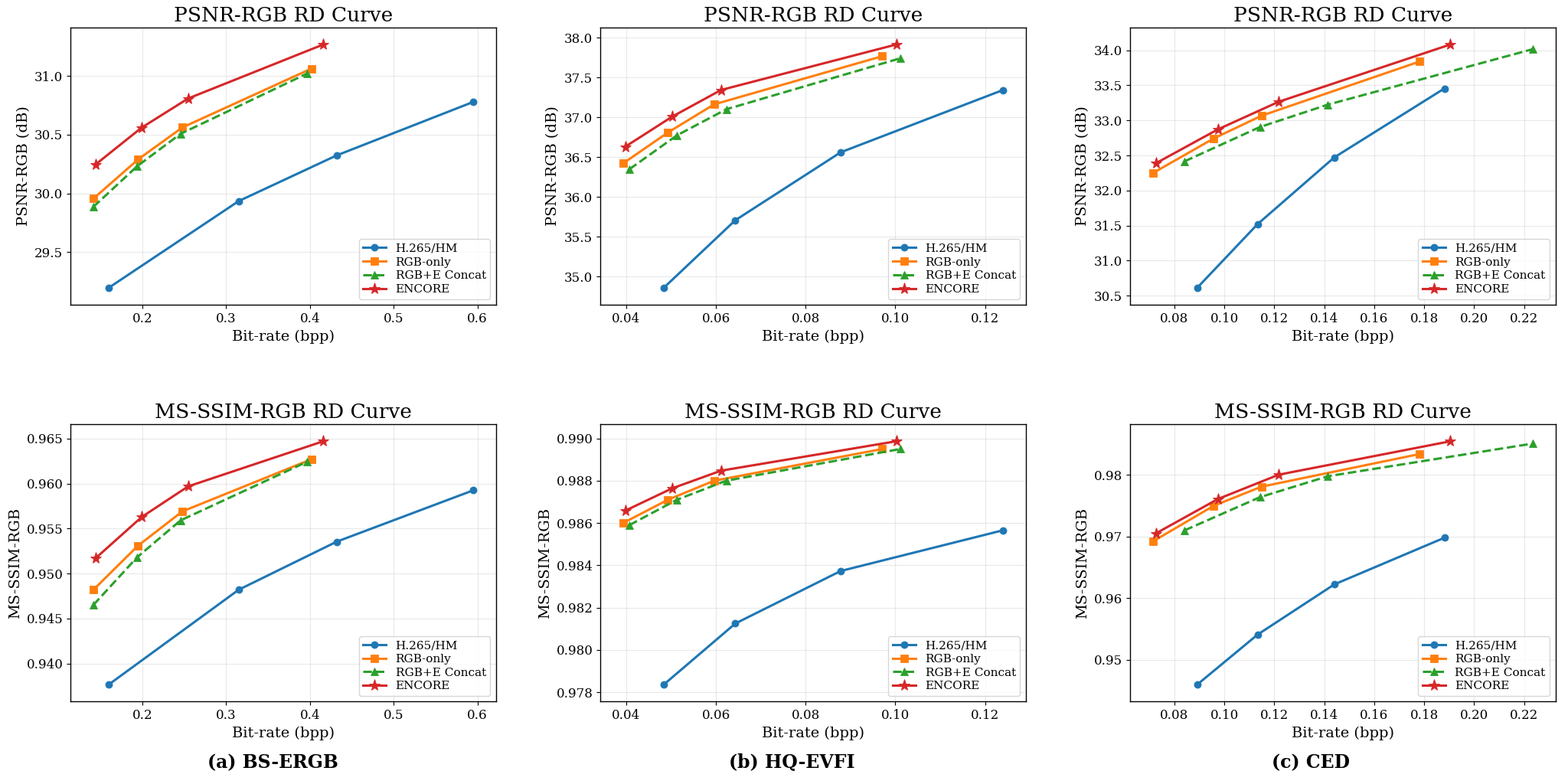}
    \caption{Rate--distortion curves on BS-ERGB, HQ-EVFI, and CED for codecs trained with GOP$=3$ and evaluated with GOP$=8$.}
    \label{fig:rd_curves}
\end{figure*}

\begin{table}[t]
\centering
\caption{Component ablation on BS-ERGB. BD-rate (\%) is relative to RGB-only (HyTIP); negative values indicate bitrate savings.}
\label{tab:module_ablation}
\scriptsize
\setlength{\tabcolsep}{3.2pt}
\renewcommand{\arraystretch}{1.10}
\resizebox{\columnwidth}{!}{%
\begin{tabular}{l || ccc | cc}
\hline
\rowcolor{tableheader}
\textbf{Method} & \textbf{CMR} & \textbf{SERIC} & \textbf{EAR} & \textbf{PSNR-RGB} & \textbf{MS-SSIM-RGB} \\
\hline\hline
RGB-only & & & & 0.00 & 0.00 \\
\rowcolor{tablerow}
RGB+E Concat & & & & +4.36 & +5.94 \\
CMR & $\checkmark$ & & & -10.85 & -9.42 \\
\rowcolor{tablerow}
CMR+SERIC & $\checkmark$ & $\checkmark$ & & -17.62 & -15.34 \\
\rowcolor{highlightcyan}
\textbf{\ENCORE} & $\checkmark$ & $\checkmark$ & $\checkmark$ & \textbf{-19.31} & \textbf{-16.73} \\
\hline
\end{tabular}%
}
\end{table}

\begin{table}[t]
\centering
\caption{Complexity comparison on BS-ERGB. GFLOPs are reported per P-frame, runtime per 8-frame ($1\mathrm{I}+7\mathrm{P}$) GOP, and BD-rate relative to RGB-only.}
\label{tab:complexity}
\scriptsize
\setlength{\tabcolsep}{1.7pt}
\renewcommand{\arraystretch}{1.10}
\resizebox{\columnwidth}{!}{%
\begin{tabular}{l || c || c || c || c || c}
\hline
\rowcolor{tableheader}
\textbf{Method} & \shortstack{\textbf{Params}\\\textbf{(M)}} & \shortstack{\textbf{Total}\\\textbf{GFLOPs}} & \shortstack{\textbf{Encoding time}\\\textbf{(ms)}} & \shortstack{\textbf{Decoding time}\\\textbf{(ms)}} & \shortstack{\textbf{PSNR-RGB}\\\textbf{BD-Rate (\%)}} \\
\hline\hline
RGB-only & 48.79 & 1989.5 & 1553.1 & 2373.1 & 0.00 \\
\rowcolor{tablerow}
RGB+E Concat & 48.98 & 2011.9 & 1642.0 & 2617.3 & +4.36 \\
\rowcolor{highlightcyan}
ENCORE (Ours) & 52.72 & 2135.0 & 1747.1 & 2576.6 & -19.31 \\
\hline
\end{tabular}
}
\end{table}

\subsection{Rate--Distortion Performance}

We compare \ENCORE with the RGB-only HyTIP backbone and a controlled direct-fusion baseline, denoted RGB+E Concat. RGB+E Concat uses the same event encoder as \ENCORE but replaces CMR, SERIC, and EAR with channel-wise feature concatenation followed by a learned projection. This design isolates the gain of complementary motion refinement from the gain obtained by merely adding event inputs. H.265/HM is additionally plotted as a conventional codec reference, but all learned-method BD-rates use RGB-only HyTIP as the anchor. Table~\ref{tab:main_results} reports the standard setting, in which learned codecs are trained with GOP$=3$ and evaluated with GOP$=8$, while Table~\ref{tab:long_gop_results} reports the long-GOP setting, in which the GOP$=3$ models are fine-tuned with GOP$=8$ and evaluated with GOP$=32$.

In the standard setting, \ENCORE improves both metrics on all three datasets. Its PSNR-RGB / MS-SSIM-RGB BD-rates are $-19.31\%/-16.73\%$ on BS-ERGB, $-9.20\%/-9.61\%$ on HQ-EVFI, and $-5.63\%/-6.49\%$ on CED. The gains on the high-frame-rate HQ-EVFI dataset and the differently captured CED dataset show that the benefit persists across varied temporal sampling and sensor conditions. In contrast, RGB+E Concat increases rate for every short-GOP comparison, by $+4.36\%$ to $+11.23\%$ in PSNR-RGB BD-rate and by $+4.57\%$ to $+9.03\%$ in MS-SSIM-RGB BD-rate. Event features differ semantically from RGB motion features and can contain substantial sensor noise and redundant activity. Direct concatenation cannot distinguish responses that benefit RGB compression or provide useful flow corrections from those that introduce irrelevant perturbations. \ENCORE addresses these limitations progressively: the motion-oriented event stem reduces the semantic gap, CMR separates shared and modality-specific motion, SERIC retains active event evidence that is novel relative to RGB, and EAR spatially controls its contribution to flow refinement.

The long-GOP setting exhibits the same trend. The PSNR-RGB / MS-SSIM-RGB BD-rates of \ENCORE are $-20.80\%/-22.14\%$ on BS-ERGB, $-9.60\%/-11.90\%$ on HQ-EVFI, and $-6.43\%/-4.35\%$ on CED. RGB+E Concat remains inconsistent: it provides a limited PSNR-RGB gain only on BS-ERGB ($-4.44\%$), but degrades at least one metric on every dataset. The larger gain on BS-ERGB under GOP$=32$ is consistent with the intended role of event-guided motion refinement, since errors in RGB-only motion estimation accumulate over a longer prediction chain.

Figure~\ref{fig:rd_curves} shows the complete short-GOP RD curves, while the long-GOP RD curves and their detailed analysis are provided in Appendix~E-A. Across both GOP settings, the \ENCORE curves consistently lie above and to the left of the RGB-only curves over the tested operating points, showing that the gains are not caused by a single quality level. RGB+E Concat is frequently below or to the right of the RGB-only reference, consistent with its positive or unstable BD-rates. The H.265/HM curves provide an additional conventional-codec reference but are not used as the BD-rate anchor. The qualitative comparisons in Fig.~\ref{fig:qualitative} provide corresponding reconstruction evidence: at comparable bitrates, \ENCORE better preserves the contours of the flame and fine structural boundaries in the indoor and outdoor crops than RGB-only and RGB+E Concat, with less visible smoothing around motion-sensitive regions. These observations are consistent with the intended effect of using events to refine temporal prediction rather than directly fusing all event responses.

\subsection{Ablation Studies}

All ablations are conducted on BS-ERGB with the same HyTIP backbone, training schedule, rate--distortion operating points, and evaluation protocol. Their BD-rates are computed relative to RGB-only, which is reported as $0.0\%$. This controlled setting attributes performance differences to the event-assisted design rather than to changes in the underlying RGB codec.

We evaluate the progressive contribution of CMR, SERIC, and EAR. RGB+E Concat provides a direct RGB--event reference, while the subsequent variants successively introduce common/specific decomposition, response calibration, and spatial routing. As shown in Table~\ref{tab:module_ablation}, RGB+E Concat degrades both metrics ($+4.36\%$ PSNR-RGB and $+5.94\%$ MS-SSIM-RGB BD-rate). CMR alone reverses this degradation, reaching $-10.85\%/-9.42\%$, which demonstrates the value of separating common and modality-specific motion rather than passing all event responses directly. Adding SERIC further improves the BD-rates to $-17.62\%/-15.34\%$, and EAR gives the final $-19.31\%/-16.73\%$. The monotonic improvement confirms that calibration and spatial routing contribute beyond the representation decomposition.

\subsection{Complexity Analysis}

All \ENCORE-specific event processing is performed within the motion-refinement branch, while the remaining HyTIP codec retains its original architecture. Table~\ref{tab:complexity} reports end-to-end complexity per P-frame and the average runtime of a real $1\mathrm{I}+7\mathrm{P}$ GOP. Relative to RGB-only, \ENCORE adds $3.93$M parameters ($8.1\%$) and $145.5$ GFLOPs per P-frame ($7.3\%$). Its measured encoding and decoding times increase by $194.0$ ms ($12.5\%$) and $203.5$ ms ($8.6\%$) per GOP, respectively. Relative to RGB+E Concat, \ENCORE adds $123.1$ GFLOPs and $105.1$ ms to encoding, while its measured decoding time is $40.7$ ms lower. In return, \ENCORE provides a $-19.31\%$ PSNR-RGB BD-rate on BS-ERGB, i.e., approximately $20\%$ bitrate savings relative to RGB-only. This trade-off demonstrates that the proposed complementary refinement yields substantial coding gains with moderate computational overhead.

\FloatBarrier
\section{Conclusion}

We presented \ENCORE, an event-assisted framework that uses events as an auxiliary modality to improve learned RGB video compression by refining, rather than replacing, RGB-estimated motion. CMR decomposes cross-modal common and modality-specific motion, SERIC identifies event-specific responses that are both active and insufficiently explained by RGB, and EAR applies the resulting correction through bounded spatial routing. Experiments on BS-ERGB, HQ-EVFI, and CED demonstrate consistent improvements across sensors and GOP lengths: under GOP$=8$ evaluation, \ENCORE achieves PSNR-RGB / MS-SSIM-RGB BD-rate savings of $19.31\%/16.73\%$, $9.20\%/9.61\%$, and $5.63\%/6.49\%$ on the three datasets, respectively, while the BS-ERGB gains reach $20.80\%/22.14\%$ with GOP$=32$. Component ablations show that RGB+E Concat can degrade compression, whereas CMR, SERIC, and EAR provide progressive improvements. These results establish event-derived complementary motion as an effective auxiliary cue for improving the rate--distortion performance of learned RGB video compression. Future work will investigate robustness to event sparsity and sensor noise and further reduce the computational overhead.

\bibliographystyle{IEEEtran}
\bibliography{references}

\clearpage
\appendices
\section{Dataset Details}

\subsection{BS-ERGB}

BS-ERGB, released with Time Lens++~\cite{tulyakov2022timelenspp}, is our primary dataset for training, evaluation, and ablation. It provides beam-splitter aligned RGB frames and events across 123 sequences, with a processed spatial resolution of approximately $970\times625$. RGB frames are captured at about 28 fps, and individual sequences contain roughly 100--600 frames. Its mixture of fast and slow motion, static and moving cameras, and non-planar scene dynamics makes it suitable for assessing whether events improve motion modeling for RGB coding.

\subsection{HQ-EVFI}

HQ-EVFI, introduced by TimeLens-XL~\cite{ma2024timelensxl}, contains 71 synchronized RGB--event sequences at $727\times602$ resolution. Its high-frame-rate recordings, diverse scenes, and large, complex motions support evaluation under different temporal sampling and motion conditions. We train and evaluate a dataset-specific model on HQ-EVFI using the same protocol as for BS-ERGB.

\subsection{CED}

CED~\cite{scheerlinck2019ced} provides a supplementary evaluation using a Color-DAVIS346 sensor. It provides color frames and color events at $346\times260$ resolution, spanning approximately 50 minutes of recordings, around 100k color frames, and more than one billion color events. Its Simple, Indoors, People, Driving, and Calibration subsets include low-light scenes, high dynamic range, rapid camera motion, and sensor characteristics that differ from the beam-splitter datasets. We train and evaluate a dataset-specific model on CED using the same protocol as for BS-ERGB and report its results separately.

\section{Evaluation Details}

For an evaluated RGB sequence with $N$ coded frames of spatial size $H\times W$, bitrate is
\begin{equation}
    \mathrm{bpp}=\frac{B_{\mathrm{total}}}{NHW},
\end{equation}
where $B_{\mathrm{total}}$ includes all coded RGB components: I-frame bits, motion latents, content latents, hyperpriors, and side information.

RGB reconstruction quality is evaluated framewise and then averaged over the coded frames. For frame $t$, we compute
\begin{align}
    \mathrm{MSE}_t&=
    \frac{1}{3HW}\left\|\mathbf{x}_t-\hat{\mathbf{x}}_t\right\|_2^2,
    \label{eq:appendix_mse}\\
    \mathrm{PSNR}_t&=
    10\log_{10}\frac{L^2}{\mathrm{MSE}_t}.
    \label{eq:appendix_psnr}
\end{align}
where $L$ is the peak RGB value. MS-SSIM-RGB~\cite{wang2003msssim} is computed in the same RGB domain. Identical crop or padding removal, I-frame handling, GOP boundaries, and frame averaging are used for RGB-only and every event-assisted variant.

Each method is evaluated at four rate--distortion operating points, yielding dataset-level bpp--PSNR-RGB and bpp--MS-SSIM-RGB curves. BD-rate~\cite{bjontegaard2001bdrate} is computed from logarithmic-rate interpolation over only the common quality range of a compared pair. For MS-SSIM, we first transform the quality value as $-10\log_{10}(1-\mathrm{MS\text{-}SSIM})$. RGB-only HyTIP is the common BD-rate anchor throughout the paper, including the ablations; a negative BD-rate denotes bitrate savings at matched RGB quality.

\section{Implementation Details}

All variants are implemented in PyTorch and initialize the RGB branch from the pretrained HyTIP model. We use Adam with $(\beta_1,\beta_2)=(0.9,0.999)$, zero weight decay, a batch size of one, and element-wise gradient clipping to $[-5,5]$. The auxiliary-loss weights in Eq.~\eqref{eq:full_loss} are $(\lambda_r,\lambda_s,\lambda_m,\lambda_w)=(0.02,0.005,0.02,0.05)$; all otherwise unspecified loss coefficients are set to $1$. Within each dataset, all learned variants use identical data splits, preprocessing, augmentation, total epoch budgets, learning rates, GOP protocols, and evaluation settings. Training follows two stages. We first freeze the pretrained RGB backbone and optimize the trainable event-assisted modules for $200$ epochs using a learning rate of $1\times10^{-4}$. We then unfreeze the RGB backbone and jointly fine-tune the entire codec for another $100$ epochs using a learning rate of $5\times10^{-5}$.

Training uses $256\times256$ random crops, horizontal flipping with probability $0.5$, and a fixed random seed of $888888$. For BS-ERGB, the official \texttt{3\_TRAINING}, \texttt{2\_VALIDATION}, and \texttt{1\_TEST} subsets are used for training, model selection, and testing, respectively. For HQ-EVFI and CED, we train separate dataset-specific models and follow the same optimization, preprocessing, GOP, and evaluation protocol as for BS-ERGB; no \ENCORE checkpoint trained on BS-ERGB is transferred to either dataset. Standard models are trained using three-frame clips (GOP$=3$) and evaluated with GOP$=8$. For long-GOP evaluation, each dataset's GOP$=3$ checkpoints are fine-tuned using GOP$=8$ with the same two-stage schedule: $200$ epochs with the RGB backbone frozen, followed by $100$ epochs of joint fine-tuning after unfreezing it. The resulting models are tested with GOP$=32$. All learned methods use the same BT.709 color conversion and four operating points, $\{21,32,42,63\}$. The complexity and runtime results in Table~\ref{tab:complexity} are measured on NVIDIA RTX PRO 6000 GPUs using BS-ERGB frames at their native $970\times625$ resolution, with internal padding to $1024\times640$ for codec processing.

\section{Theoretical Analysis of Calibration and Routing}

This appendix provides idealized analyses of the cross-modal novelty, multiplicative calibration, and bounded residual routing used by \ENCORE. The results explain useful structural properties of the design; they do not imply that the learned projection, correction predictor, or routing network always attains its corresponding oracle.

\subsection{Innovation Interpretation of Cross-Modal Novelty}

Let $\mathbf{X}=\mathbf{F}_t^r$ and $\mathbf{Y}=\mathbf{F}_t^{e,s}$ denote square-integrable aligned feature vectors, and consider an RGB-to-event predictor optimized under mean-squared error. Its population-optimal solution is
\begin{equation}
    \psi^*(\mathbf{X})=\mathbb{E}[\mathbf{Y}\mid\mathbf{X}].
\end{equation}
Define the event innovation
\begin{equation}
    \boldsymbol{\epsilon}
    =\mathbf{Y}-\mathbb{E}[\mathbf{Y}\mid\mathbf{X}].
    \label{eq:event_innovation}
\end{equation}

\noindent\textbf{Lemma 1 (Orthogonality of event innovation).}
For every square-integrable function $\mathbf{h}(\mathbf{X})$ of compatible dimension,
\begin{equation}
    \mathbb{E}\!\left[\boldsymbol{\epsilon}^{\mathsf T}\mathbf{h}(\mathbf{X})\right]=0.
    \label{eq:innovation_orthogonality}
\end{equation}
In particular, $\mathbb{E}[\boldsymbol{\epsilon}^{\mathsf T}\mathbf{X}]=0$ whenever the dimensions are compatible.

\noindent\emph{Proof.}
By the law of iterated expectation and because $\mathbf{h}(\mathbf{X})$ is fixed conditional on $\mathbf{X}$,
\begin{align}
\mathbb{E}\!\left[\boldsymbol{\epsilon}^{\mathsf T}\mathbf{h}(\mathbf{X})\right]
&=\mathbb{E}\!\left[
\mathbb{E}\!\left[
(\mathbf{Y}-\mathbb{E}[\mathbf{Y}\mid\mathbf{X}])^{\mathsf T}
\mathbf{h}(\mathbf{X})\mid\mathbf{X}
\right]\right] \\
&=0.
\end{align}

\noindent\textbf{Lemma 2 (Energy of the unexplained component).}
The expected squared innovation is
\begin{equation}
    \mathbb{E}\!\left[\|\boldsymbol{\epsilon}\|_2^2\right]
    =
    \mathbb{E}\!\left[
    \operatorname{tr}\!\left(\operatorname{Cov}(\mathbf{Y}\mid\mathbf{X})\right)
    \right].
    \label{eq:innovation_energy}
\end{equation}

\noindent\emph{Proof.}
Given $\mathbf{X}$, $\boldsymbol{\epsilon}$ has zero mean and covariance $\operatorname{Cov}(\mathbf{Y}\mid\mathbf{X})$:
\begin{equation}
\mathbb{E}[\|\boldsymbol{\epsilon}\|_2^2\mid\mathbf{X}]
=\operatorname{tr}(\operatorname{Cov}(\mathbf{Y}\mid\mathbf{X})),
\end{equation}
and taking expectation over $\mathbf{X}$ proves Eq.~\eqref{eq:innovation_energy}.

The decomposition $\mathbf{Y}=\mathbb{E}[\mathbf{Y}\mid\mathbf{X}]+\boldsymbol{\epsilon}$ separates the RGB-predictable component from event variation unexplained by RGB. Therefore, a small residual suggests redundancy with the RGB representation, whereas a large residual indicates potentially complementary dynamics. Equation~\eqref{eq:event_novelty} uses the absolute discrepancy followed by local aggregation, rather than the squared population residual assumed above. It can therefore be interpreted as a robust local surrogate for innovation magnitude. The learned $\psi$ need not equal the conditional-mean predictor, so this interpretation is structural, not a guarantee.

\subsection{Non-Expansiveness of SERIC Calibration}

For the following results, tensors are viewed as vectors after flattening spatial and channel dimensions; spatial utility values are broadcast over channels as in Eq.~\eqref{eq:calibrated_event}.

\noindent\textbf{Lemma 3 (Non-expansive multiplicative calibration).}
For every $p\in[1,\infty]$,
\begin{equation}
    \|\widetilde{\mathbf{F}}_t^{e,s}\|_p
    =\|\mathbf{U}_t\odot\mathbf{F}_t^{e,s}\|_p
    \leq\|\mathbf{F}_t^{e,s}\|_p.
    \label{eq:seric_nonexpansive}
\end{equation}

\noindent\emph{Proof.}
Since $0\leq U_{t,i}\leq1$, every element satisfies
$|U_{t,i}F_{t,i}^{e,s}|\leq|F_{t,i}^{e,s}|$. Summing the $p$th powers proves the result for finite $p$, and taking the maximum proves it for $p=\infty$.

Consequently, the multiplicative calibration alone cannot amplify weak, redundant, or noisy event responses; it can only preserve or attenuate them.

\noindent\textbf{Proposition 1 (Bound on event-induced flow perturbation).}
Fix $\mathbf{F}_t^c$ and assume that $\mathcal{R}$ is $L_{\mathcal{R}}$-Lipschitz continuous with respect to its event-feature argument. Then
\begin{align}
&\left\|
\mathcal{R}([\mathbf{F}_t^c,\widetilde{\mathbf{F}}_t^{e,s}])
-\mathcal{R}([\mathbf{F}_t^c,\mathbf{0}])
\right\|_p \\
&\qquad\leq L_{\mathcal{R}}\|\widetilde{\mathbf{F}}_t^{e,s}\|_p
\leq L_{\mathcal{R}}\|\mathbf{F}_t^{e,s}\|_p.
\label{eq:event_flow_bound}
\end{align}

\noindent\emph{Proof.}
The first inequality follows from the Lipschitz assumption after setting the two event arguments to $\widetilde{\mathbf{F}}_t^{e,s}$ and $\mathbf{0}$. The second follows directly from Lemma~3.

This proposition bounds the change in the candidate correction attributable to the calibrated event branch. It does not prove that SERIC always retains the correct evidence; it establishes only non-amplification of feature norm and, under the stated regularity assumption, a corresponding bound on output perturbation.

\subsection{Optimality and Stability of Energy-Aware Routing}

Analyze Eq.~\eqref{eq:routed_refinement} independently at one spatial location. Let $\mathbf{v}^*\in\mathbb{R}^2$ be the ideal displacement, $\mathbf{v}^r\in\mathbb{R}^2$ the RGB estimate, $\Delta\mathbf{v}\in\mathbb{R}^2$ the event-derived candidate correction, and $g\in[0,1]$ the routing weight shared by the horizontal and vertical components. Define
\begin{equation}
    \phi(g)=\|\mathbf{v}^*-(\mathbf{v}^r+g\Delta\mathbf{v})\|_2^2.
    \label{eq:local_routing_error}
\end{equation}

\noindent\textbf{Proposition 2 (Oracle optimality of bounded residual routing).}
If $\Delta\mathbf{v}\neq\mathbf{0}$, the minimizer of Eq.~\eqref{eq:local_routing_error} over $g\in[0,1]$ is
\begin{equation}
    g^*
    =\Pi_{[0,1]}\!\left(
    \frac{\langle\mathbf{v}^*-\mathbf{v}^r,\Delta\mathbf{v}\rangle}
    {\|\Delta\mathbf{v}\|_2^2}
    \right),
    \label{eq:oracle_gate}
\end{equation}
where $\Pi_{[0,1]}$ projects onto $[0,1]$. Consequently,
\begin{equation}
    \phi(g^*)\leq\min\{\phi(0),\phi(1)\}.
    \label{eq:oracle_endpoint_bound}
\end{equation}

\noindent\emph{Proof.}
Let $\mathbf{e}=\mathbf{v}^*-\mathbf{v}^r$. Expanding gives
\begin{equation}
    \phi(g)=\|\mathbf{e}\|_2^2
    -2g\langle\mathbf{e},\Delta\mathbf{v}\rangle
    +g^2\|\Delta\mathbf{v}\|_2^2.
\end{equation}
This is a convex quadratic whose unconstrained minimizer is
$\langle\mathbf{e},\Delta\mathbf{v}\rangle/\|\Delta\mathbf{v}\|_2^2$.
Projecting it onto the feasible interval yields Eq.~\eqref{eq:oracle_gate}. Since $0$ and $1$ are feasible, the minimizer cannot have larger error than either endpoint.

The endpoints have direct interpretations: $g=0$ rejects the event correction and recovers the RGB flow, whereas $g=1$ applies the full candidate correction. Geometrically, the oracle rejects a correction when $\langle\mathbf{v}^*-\mathbf{v}^r,\Delta\mathbf{v}\rangle\leq0$, uses a partial correction when this inner product lies strictly between $0$ and $\|\Delta\mathbf{v}\|_2^2$, and applies the full correction when it is at least $\|\Delta\mathbf{v}\|_2^2$. A continuous routing value is therefore more flexible than a binary accept/reject decision. EAR should be interpreted as a learned approximation to this spatial oracle, not as a guarantee that $\mathbf{G}_t=g^*$.

\noindent\textbf{Corollary 1 (Stability of bounded residual routing).}
For every $p\in[1,\infty]$,
\begin{equation}
    \|\widetilde{\mathbf{v}}_t-\mathbf{v}_t^r\|_p
    =\|\mathbf{G}_t\odot\Delta\mathbf{v}_t\|_p
    \leq\|\Delta\mathbf{v}_t\|_p.
    \label{eq:routing_stability}
\end{equation}

\noindent\emph{Proof.}
The result follows element-wise from $0\leq G_{t,i}\leq1$ and by the same finite-$p$ and $p=\infty$ arguments used in Lemma~3.

Thus, EAR cannot amplify the magnitude of the candidate correction; it preserves or attenuates the proposed update while retaining the RGB flow as the residual reference. Together, the analyses above clarify the progressive design: the cross-modal discrepancy approximates event innovation relative to RGB, SERIC applies a non-expansive filter to this complementary evidence, and EAR performs a bounded residual update whose oracle is at least as accurate as rejecting or fully accepting the candidate correction.

\section{Supplementary Experimental Results}

\subsection{Long-GOP Rate--Distortion Curves}

\begin{figure*}[ht]
    \centering
    \includegraphics[width=\textwidth]{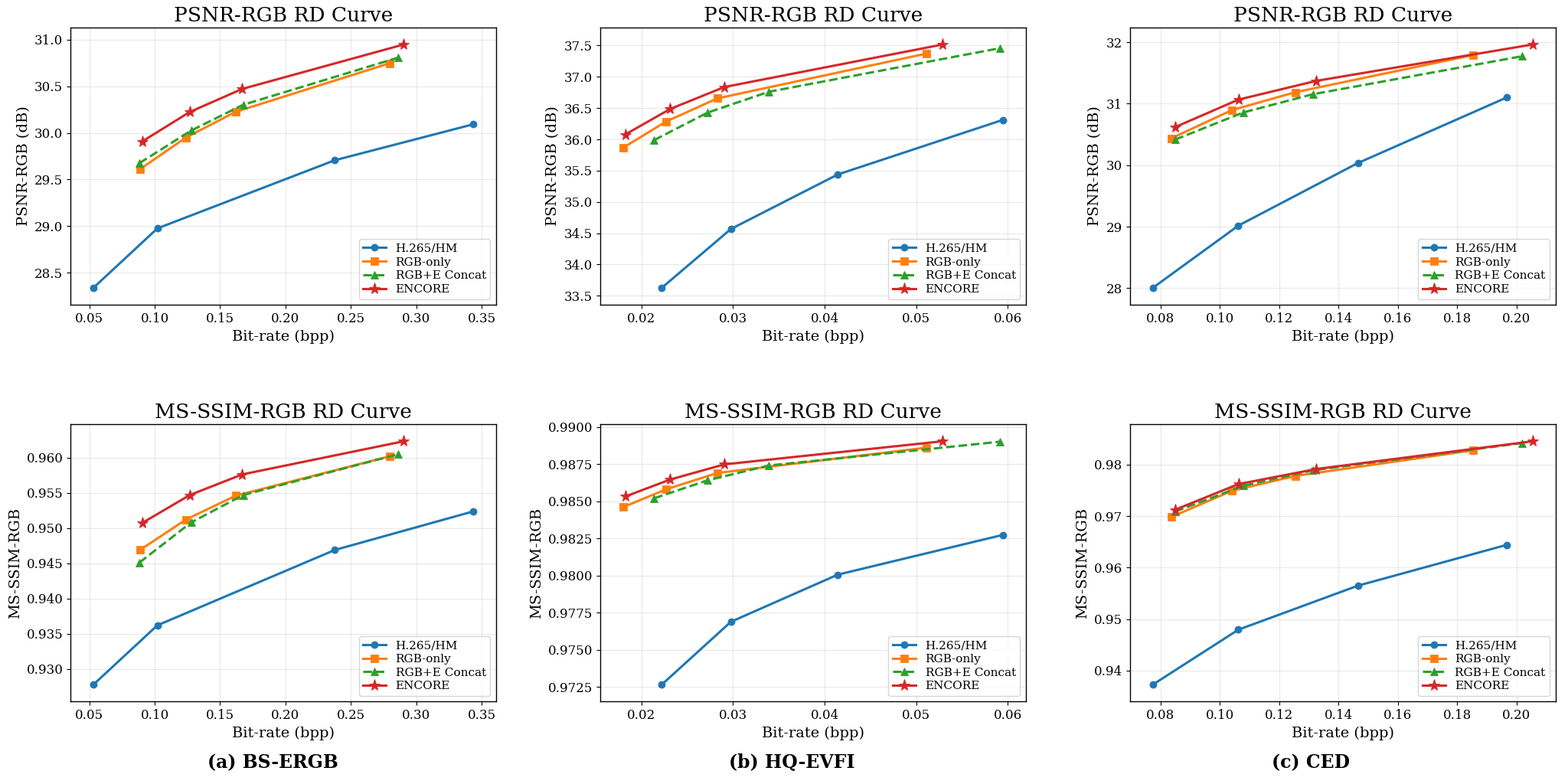}
    \caption{Long-GOP rate--distortion curves on BS-ERGB, HQ-EVFI, and CED for codecs initialized from GOP$=3$, fine-tuned with GOP$=8$, and evaluated with GOP$=32$.}
    \label{fig:rd_curves_long_gop}
\end{figure*}

Figure~\ref{fig:rd_curves_long_gop} confirms that the benefit of event-guided refinement persists over the longer prediction chain. Over the evaluated rate range, the \ENCORE curves remain consistently above or to the left of the RGB-only curves for both PSNR-RGB and MS-SSIM-RGB. Relative to RGB-only, \ENCORE achieves PSNR-RGB / MS-SSIM-RGB BD-rate savings of $-20.80\%/-22.14\%$ on BS-ERGB, $-9.60\%/-11.90\%$ on HQ-EVFI, and $-6.43\%/-4.35\%$ on CED. The improvement is strongest on BS-ERGB, where motion-estimation errors can accumulate within the longer GOP$=32$ prediction chain. By contrast, RGB+E Concat remains unstable: it improves PSNR-RGB only on BS-ERGB but degrades at least one metric on every dataset. These results indicate that longer temporal propagation does not make unfiltered event fusion reliably useful; the separation, calibration, and routing stages remain necessary for consistent coding gains.

\end{document}